\documentclass{article}

\usepackage{microtype}
\usepackage{graphicx}
\usepackage{subcaption}
\usepackage{booktabs} 
\usepackage{colortbl}
\usepackage{xcolor}
\usepackage{booktabs}
\usepackage{array}

\usepackage{hyperref}

\usepackage[accepted]{icml2026}

\usepackage{amsmath}
\usepackage{amssymb}
\usepackage{mathtools}
\usepackage{amsthm}
\usepackage{enumitem}

\usepackage[capitalize,noabbrev]{cleveref}

\theoremstyle{plain}

\theoremstyle{definition}

\theoremstyle{remark}

\usepackage[textsize=tiny]{todonotes}

\icmltitlerunning{FirstPass: Grounding AI Scientific Judgment in Multi-Round Editorial Outcomes}

\begin{document}

\twocolumn[
  \icmltitle{FirstPass: Grounding AI Scientific Judgment in Multi-Round Editorial Outcomes}



  \icmlsetsymbol{equal}{*}

  \begin{icmlauthorlist}
    \icmlauthor{Prabhjot Singh}{aut,rdi}
    \icmlauthor{Somnath Luitel}{wku}
    \icmlauthor{Manmeet Singh}{wku}
    \icmlauthor{Josh Durkee}{wku}
  \end{icmlauthorlist}

  \icmlaffiliation{aut}{The University of Texas at Austin, Austin, TX, USA}
  \icmlaffiliation{rdi}{RediMinds Inc., USA}
  \icmlaffiliation{wku}{Disaster Science Operations Center, Western Kentucky University, Bowling Green, KY, USA}

  \icmlcorrespondingauthor{Prabhjot Singh}{prabhjot.singh@utexas.edu}

  \icmlkeywords{scientific peer review, multi-round dialogue, loss masking, editorial outcome prediction, domain generalization, AI co-author, fine-tuning, Nature Communications, revision-cycle prediction, outcome-grounded evaluation}

  \vskip 0.3in
]



\printAffiliationsAndNotice{}  

\begin{abstract}
AI systems for peer review fail on three fronts: they train on Computer Science and Machine Learning venues alone, ignore the iterative dialogue that validates science, and evaluate on stylistic mimicry rather than real editorial judgment. We introduce \textsc{FirstPass}, a dataset and fine-tuned model that addresses all three. Curating 3,668 complete multi-round peer-review dialogues from \textit{Nature Communications} across five scientific domains (biology, chemistry, neuroscience, physics, and earth science), we exploit mandatory transparent peer review (instituted November 2022) and verify 100\% content integrity by automated audit. We fine-tune Qwen2.5-7B-Instruct via Low-Rank Adaptation (LoRA) on three tasks: review generation, reviewer updating, and revision-cycle prediction. Our key finding is that response-only loss masking is a prerequisite, not an optimization: without it, accuracy is 62.0\%, below the majority baseline; with it, \textsc{FirstPass} achieves 80.5\% accuracy and F1-macro 78.2\% on predicting editorial outcomes (\textsc{Standard} vs.\ \textsc{Extended} revision cycles), outperforming Gemini-3.1-flash-lite-preview zero-shot by 10.4 percentage points and all baselines with statistical significance (McNemar $p < 0.001$). On generation, \textsc{FirstPass} produces reviews averaging 1,187 words, substantially closer to human references (2,155 words) than any baseline, achieving ROUGE-L 0.154 with significant gains over Qwen and DeepSeek zero-shot ($p < 0.001$). Deployed in the pre-submission loop as an anticipatory scientific co-author, \textsc{FirstPass} simulates expert critique and predicts revision cycle outcomes \emph{before} submission, giving authors the judgment a trusted colleague would provide, with consistent cross-domain performance across five disciplines.
\end{abstract}

\section{Introduction}

Scientific peer review is collapsing under its own success. Submission volumes at high-impact journals have doubled in five years; reviewer pools have not. The result is delayed discovery, burned-out experts, and declining review quality, precisely when accelerating climate science, pandemic preparedness, and materials discovery demand faster validation.

Large Language Models offer a tempting fix, but current AI review systems fail on three fronts. \textbf{Domain narrowness:} Every major dataset, PeerRead~\cite{kang2018peerread}, ReviewMT~\cite{tan2025reviewmt}, and MARG~\cite{darcy2024marg}, draws exclusively from CS/ML conferences. A model trained on ICLR reviews learns to critique ablation studies; it has never seen a biology reviewer demand contamination controls or a chemist question NMR spectral assignments. \textbf{Static treatment:} Peer review is dialogue, not monologue. Existing systems generate reviews in one shot, blind to the author-response iterations where scientific claims are actually stress-tested. \textbf{Circular evaluation:} Systems are graded on whether they look like good reviews, not whether they align with what editors actually demanded be changed. These three failures share a common root: they treat peer review as a one-shot text generation task rather than as an exercise in scientific judgment. A trusted co-author does not merely produce a review-shaped paragraph. They tell you which methodological concerns will survive the rebuttal, which claims a domain expert will challenge, and whether your manuscript will require a second revision cycle. No existing system has been trained or evaluated to perform this judgment. \textsc{FirstPass} is.

We introduce \textsc{FirstPass}\renewcommand{\thefootnote}{\fnsymbol{footnote}}\setcounter{footnote}{1}\footnote{Code, dataset, and model weight: \url{https://github.com/prabhjotschugh/firstpass-peer-review}.}\renewcommand{\thefootnote}{\arabic{footnote}}, the first AI review system trained on complete multi-round peer-review dialogues across five scientific domains, with evaluation grounded in real editorial outcomes. Our central finding reshapes how to train LLMs on long scientific documents: response-only loss masking is a prerequisite, not an optimization. Without it, accuracy collapses to 62.0\%, below the majority baseline. With it, \textsc{FirstPass} achieves 80.5\% accuracy and F1-macro 78.2\% on revision-cycle prediction, outperforming Gemini-3.1-flash-lite-preview zero-shot by 10.4 percentage points ($p < 0.001$, McNemar's exact test). The practical consequence is direct. An author who submits without anticipatory critique learns what expert reviewers demand only after submission, when the rebuttal clock is ticking and revision cycles are costly. \textsc{FirstPass} closes this loop upstream: trained on complete editorial dialogues, it generates simulated expert reviews and predicts revision-cycle outcomes with state-of-the-art accuracy across five scientific disciplines. This framing is a deployment hypothesis - our evaluation measures prediction accuracy on completed dialogues, and prospective validation in which authors use \textsc{FirstPass} pre-submission and subsequent outcomes are tracked against predictions remains the most direct path to confirming the co-authorship claim. This is precisely the judgment that defines the tool-to-co-author boundary at the heart of the \textit{AI Scientists: Tools, Co-authors, or Founders?} workshop at ICML 2026.\footnote{\url{https://ai4sciencecommunity.github.io/icml26}} Our contributions:
\begin{enumerate}[itemsep=3pt, parsep=0pt, topsep=3pt]
    \item \textbf{FirstPass dataset:} 3,668 multi-domain, multi-round peer-review dialogues from \textit{Nature Communications} with 100\% verified content integrity.
    \item \textbf{Outcome-grounded evaluation:} predicting real editorial decisions with 80.5\% accuracy and F1-macro 78.2\% across five scientific domains.
    \item \textbf{The masking finding:} empirical proof that response-only loss masking is critical for long-context scientific classification, with an 18.5 percentage point swing between masked and unmasked variants.
    \item \textbf{A pre-submission co-authorship use case:} \textsc{FirstPass} as an anticipatory reviewer that simulates expert critique and predicts revision cycle outcomes before submission, enabling authors to strengthen manuscripts and shorten rebuttal cycles.
\end{enumerate}

\section{Related Work}

\textbf{Peer review datasets.} PeerRead~\cite{kang2018peerread} established the CS/ML-only precedent: 14.7K drafts from ACL, NIPS, and ICLR. ReviewMT~\cite{tan2025reviewmt} added multi-turn dialogue structure and MARG~\cite{darcy2024marg} introduced multi-agent generation, but both remain anchored to ML venues and neither evaluates against real editorial outcomes. A small \textit{Nature Communications} sample appears in ReviewMT but is not the basis for training or evaluation. \textsc{FirstPass} is the first dataset built primarily on a multidisciplinary high-impact journal, covering five non-ML scientific domains, with outcome labels derived from actual editorial decisions rather than human ratings of generated text. \textsc{FirstPass} addresses all three gaps simultaneously: a multi-domain dataset from a high-impact natural science journal, training on the complete multi-round dialogue, and evaluation against real editorial decisions rather than stylistic proxies.

\textbf{LLM-assisted review.} Liang et al.~\cite{liang2024can} demonstrated that GPT-4 feedback matches human-human agreement rates in CS, but this finding does not transfer to natural sciences where methodological norms differ fundamentally: biology reviewers assess experimental controls and causal claim strength; chemistry reviewers interrogate spectroscopic characterization and synthesis reproducibility; neither appears in ML training corpora. The AI Scientist~\cite{lu2024aiscientist} automates the full research lifecycle including review, but remains confined to ML. Crucially, no existing system trains on the complete author-reviewer dialogue, the iterative exchange where scientific claims are genuinely stress-tested and reviewer assessments updated.

\textbf{Scientific fine-tuning and loss masking.} Recent benchmarking establishes Qwen2.5-7B-Instruct as the strongest 7B-scale model for scientific reasoning and the largest beneficiary of domain-specific fine-tuning across multi-discipline benchmarks~\cite{wang2026chartingempiricallawsllm}, directly motivating our base model choice. Response-only loss masking during instruction fine-tuning, implemented via Unsloth~\cite{unsloth2024}, has been adopted in recent alignment work but its role in long-input/short-output classification tasks, where thousands of input tokens dwarf a one-word target label, has not been empirically characterised. \textsc{FirstPass} provides the first controlled ablation demonstrating that omitting masking in this regime is not a minor degradation but a catastrophic one, dropping accuracy below the majority baseline. 

\section{The \textsc{FirstPass} Dataset}

\begin{figure}[h]
\centering
\includegraphics[width=\columnwidth]{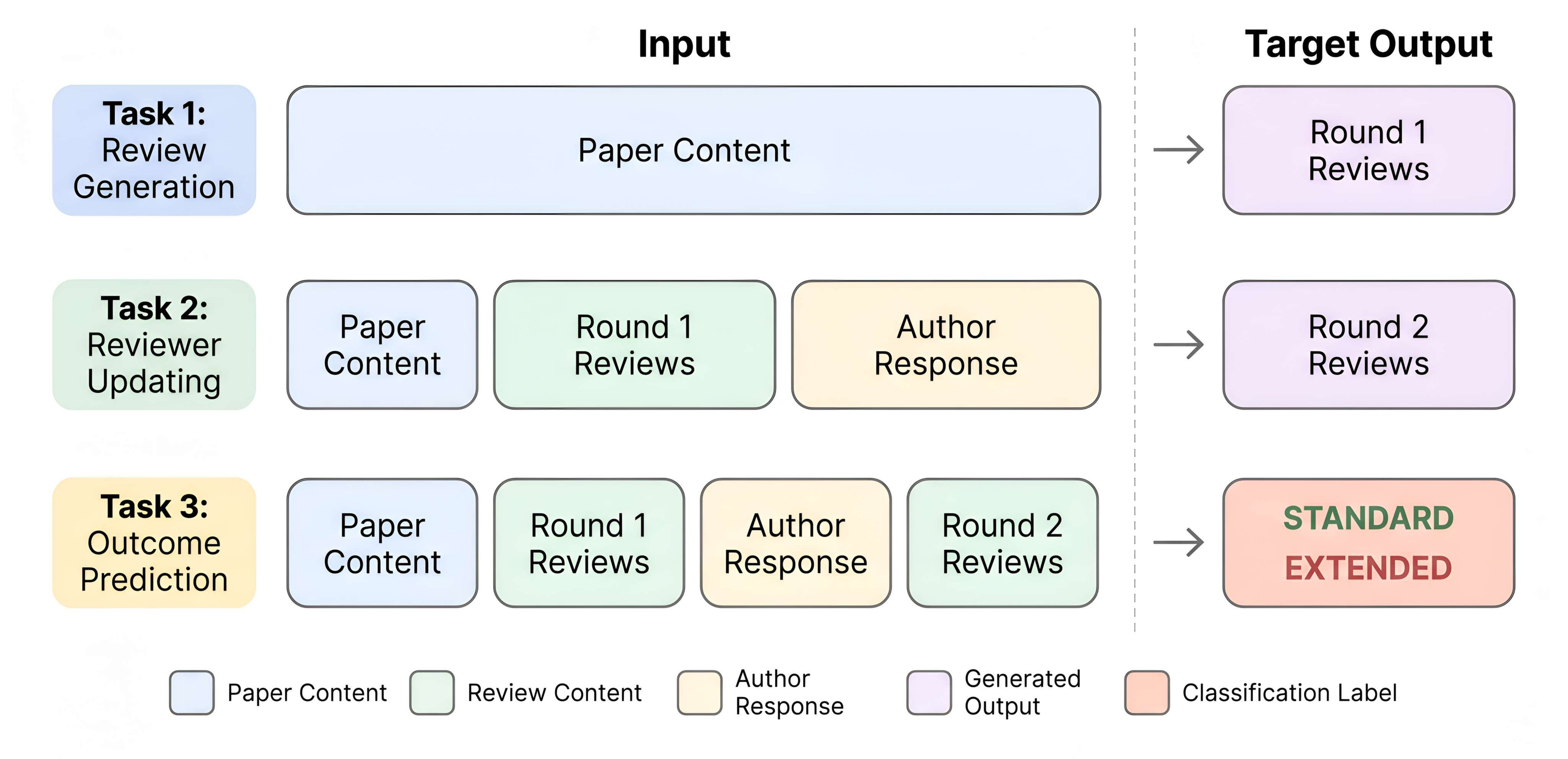}
\caption{The \textsc{FirstPass} three-task training curriculum. Each paper generates up to three examples. Task 3 (outcome prediction) is the primary evaluation task; Tasks 1 and 2 provide auxiliary training signal for review generation and reviewer updating.}
\label{fig:pipeline}
\end{figure}

The design of \textsc{FirstPass} rests on a specific hypothesis: scientific judgment is expressed not in a single review, but in the \emph{trajectory} of a dialogue. A reviewer who demands contamination controls in Round~1 and accepts the author's corrected experiment in Round~2 has modeled the scientific argument and updated accordingly. A reviewer who repeats the same concern across rounds despite author responses signals unresolved methodological debt that the editor will ultimately act on. These patterns are invisible in single-round review data. They are precisely the signal that outcome-grounded, multi-round training is designed to capture, and they separate a system that mimics reviewer prose from one that exercises genuine scientific judgment.

\textbf{Source.} We build on \textit{Nature Communications} for four reasons: (1) mandatory transparent peer review since November 2022 eliminates opt-in bias; (2) multidisciplinary scope provides breadth no CS/ML dataset offers; (3) CC BY 4.0 licensing permits derivative use; (4) reviews average 2,155 words, substantially denser than conference reviews.

\textbf{Collection.} We query the Springer Nature OpenAccess API (ISSN 2041-1723) for papers published January 2023 to December 2025, scrape article landing pages for PDFs, and parse both paper and peer-review PDFs through Gemini-3.1-flash-lite-preview~\cite{google2024gemini} using engineered extraction prompts with six-layer JSON recovery~\cite{jsonrepair2024}. A record is retained only if all four sections (abstract, introduction, methods, results) exceed 20 words and at least two complete review rounds are present.

\textbf{Integrity.} Automated audit: 3,668 records, zero hollow files, 100\% content integrity. To characterise model-in-the-loop noise introduced by Gemini-based PDF parsing, we manually verified a random sample of 60 records (approximately 1.6\% of the corpus) against source PDFs, finding a field-level error rate of 2.1\% concentrated in reference list formatting and mid-sentence line-break artefacts; no errors affected abstract, methods, results, or reviewer dialogue content, confirming that label integrity is uncompromised. Domain distribution: biology (741), chemistry (744), neuroscience (739), physics (727), earth science (717).

\textbf{Labels and tasks.} Outcome labels derive from round count: two rounds $=$ \textsc{Standard}, three or more $=$ \textsc{Extended}. This reflects editorial assessment of outstanding concerns without dependence on decision letter text, absent in 97.7\% of records. Three training examples are constructed per paper, as illustrated in Figure~\ref{fig:pipeline}: (1) \textit{review generation} (paper $\rightarrow$ Round 1 reviews); (2) \textit{reviewer updating} (paper + Round 1 + author response $\rightarrow$ Round 2 reviews); (3) \textit{outcome prediction} (full dialogue $\rightarrow$ label). Paper-level 80/10/10 split, stratified by domain and label, prevents leakage. Test set: 318 classification examples, 372 generation examples.

\section{Method}

\textbf{Base model.} We use Qwen2.5-7B-Instruct~\cite{yang2025qwen} as our foundation. Its 32,768-token context window accommodates full peer-review dialogues that routinely exceed 10,000 tokens. Benchmarking shows it achieves state-of-the-art performance at the 7B scale on scientific reasoning and long-context instruction following, and yields the largest fine-tuning gains among 7B models on multi-discipline scientific benchmarks~\cite{wang2026chartingempiricallawsllm}, critical for transferring across biology, chemistry, physics, neuroscience, and earth science.

\textbf{LoRA configuration.} We apply Low-Rank Adaptation~\cite{hu2022lora} with rank $r=32$, scaling parameter $\alpha=64$, and dropout $0.0$. We target all seven projection matrices: query, key, value, output (attention), and gate, up, down (MLP). This broader targeting outperforms attention-only LoRA in preliminary experiments. We use rank-stabilized LoRA scaling (rsLoRA) and implement via Unsloth~\cite{unsloth2024}, which provides approximately $2\times$ training speed improvement and 60\% memory reduction through custom CUDA kernels and optimized gradient checkpointing. Trainable parameters constitute approximately 2.7\% of the full 7B parameter count.

\textbf{Response-only loss masking.} This is the single most consequential design decision. Standard instruction fine-tuning computes cross-entropy loss over the complete token sequence, including thousands of input tokens the model has already seen as context. For our classification task, where inputs routinely exceed 10,000 tokens (full paper plus multi-round dialogue) and the target output is a single word (\textsc{Standard} or \textsc{Extended}), this is catastrophic: gradient updates are dominated by input token prediction, and the classification signal is effectively drowned out. The model learns to predict paper text it has already seen rather than the editorial outcome.

We apply \texttt{train\_on\_responses\_only()} from Unsloth, which identifies assistant turn boundaries via chat template markers (\texttt{<|im\_start|>assistant\textbackslash{}n} for Qwen) and sets all non-assistant token positions to label $= -100$, excluding them from loss computation. The ablation result is unambiguous and dramatic: without masking, accuracy collapses to 62.0\%, below even the 65.4\% majority baseline; with masking, accuracy reaches 80.5\%. This 18.5 percentage point swing, taking the model from worse-than-trivial to state-of-the-art, confirms that masking is not a hyperparameter optimization but an architectural prerequisite for this task regime. This finding extends beyond peer review: any long-input/short-output classification task in which document tokens vastly outnumber the target label faces the same gradient drowning problem. Response-only masking is the correct default for this entire class of tasks; \textsc{FirstPass} is the first controlled empirical demonstration at scale.

\textbf{Training configuration.} We train two separate LoRA adapters to prevent task interference.

\textit{Classification adapter (CLS):} Trains exclusively on outcome prediction examples for 3 epochs. Maximum sequence length: 12,288 tokens. Per-device batch size: 2. Gradient accumulation steps: 8 (effective batch size: 16). Learning rate: $5 \times 10^{-5}$. Scheduler: cosine with 30 warmup steps. Optimizer: \texttt{paged\_adamw\_8bit}. Precision: bfloat16. We train for 3 epochs because classification converges slowly on this imbalanced binary task (65.4\% \textsc{Standard}).

\textit{Generation adapter (SFT):} Trains jointly on review generation and reviewer updating examples for 1 epoch only. Maximum sequence length: 16,384 tokens. Identical hyperparameters otherwise. We use a single epoch because generation tasks have substantially more examples and are prone to overfitting on stylistic n-grams and repetitive phrasing with extended training.

Both adapters train on an NVIDIA GH200 120GB GPU. We select the best checkpoint by validation loss using \texttt{load\_best\_model\_at\_end}. Training completes in approximately 4 to 6 hours (CLS) and 8 to 12 hours (SFT).

\textbf{Truncation strategy.} Inputs exceeding maximum sequence length are truncated symmetrically: the first 55\% of tokens is retained (preserving paper abstract, introduction, methods, and early results) and the last 45\% is retained (preserving the most recent dialogue turns and Round 2 reviews), with a \texttt{[... content truncated ...]} marker inserted at the boundary. This ensures the model always sees both the paper's scientific content and the most recent reviewer exchange, which are the most informative signals for assessing whether concerns have been resolved. Ablations in preliminary experiments showed this outperforms simple head or tail truncation.

\textbf{Inference.} For classification, we use greedy decoding (\texttt{do\_sample=False}, \texttt{temperature=None}, \texttt{top\_p=None}) with \texttt{max\_new\_tokens=16}. The predicted label is extracted by scanning generated text for \textsc{Standard} or \textsc{Extended}, checking the final line first, then all lines, with fallback to the majority label if extraction fails. For generation, we use greedy decoding with \texttt{max\_new\_tokens=1500} and a repetition penalty of 1.1 to reduce degenerate repetition in long-form review outputs.

\section{Experiments}

\subsection{Revision-Cycle Prediction}

We evaluate eight systems on 318 test examples across all five domains (Table~\ref{tab:cls}). The majority baseline always predicts \textsc{Standard}. Zero-shot and few-shot baselines use identical system prompts. API baselines use Llama-3-8B-Instruct, DeepSeek-R1-Distill-Qwen-7B (HuggingFace router), and Gemini-3.1-flash-lite-preview (Google API), all at temperature 0. Statistical significance via McNemar's exact test.

\definecolor{firstpassgray}{RGB}{20, 100, 145}
\definecolor{headerblue}{RGB}{219, 234, 254}

\begin{table}[h]
\centering
\resizebox{\columnwidth}{!}{%
\begin{tabular}{>{\raggedright\arraybackslash}p{3.8cm} ccc}
\toprule
\rowcolor{headerblue}
\textbf{Model} & \textbf{F1-mac (\%)} & \textbf{F1-EXT (\%)} & \textbf{McNemar $p$} \\
\midrule
Majority baseline              & 39.5 \scriptsize{[37.5, 41.3]} & 0.0  & $<$0.001 \\[2pt]
Qwen2.5-7B zero-shot           & 73.3 \scriptsize{[67.6, 78.2]} & 63.3 & $0.185^{\dagger}$ \\[2pt]
Qwen2.5-7B 5-shot              & 39.5 \scriptsize{[37.5, 41.4]} & 0.0  & $<$0.001 \\[2pt]
Llama-3-8B zero-shot           & 48.3 \scriptsize{[43.1, 53.6]} & 16.4 & $<$0.001 \\[2pt]
DeepSeek-R1-7B zero-shot       & 39.5 \scriptsize{[37.5, 41.3]} & 0.0  & $<$0.001 \\[2pt]
Gemini-3.1-flash-lite-preview ZS       & 56.3 \scriptsize{[50.1, 62.1]} & 31.7 & $<$0.001 \\[2pt]
Qwen2.5-7B + LoRA (no masking) & 56.2 \scriptsize{[50.9, 61.6]} & 40.4 & $<$0.001 \\[2pt]
\textcolor{firstpassgray}{\textbf{\textsc{FirstPass} (ours)}}
  & \textcolor{firstpassgray}{\textbf{78.2}} \scriptsize{[73.1, 82.8]}
  & \textcolor{firstpassgray}{\textbf{71.0}}
  & \textcolor{firstpassgray}{\textbf{---}} \\
\bottomrule
\end{tabular}}
\vspace{10pt}
\caption{Revision-cycle prediction results ($n=318$). F1-EXT = minority class F1 (\textsc{Extended}). Bootstrap 95\% CIs in brackets. Accuracy visualised in Figure~\ref{fig:accuracy}. $^\dagger$Not significant vs.\ \textsc{FirstPass}.}
\label{tab:cls}
\end{table}

\begin{figure}[h]
\centering
\includegraphics[width=\columnwidth]{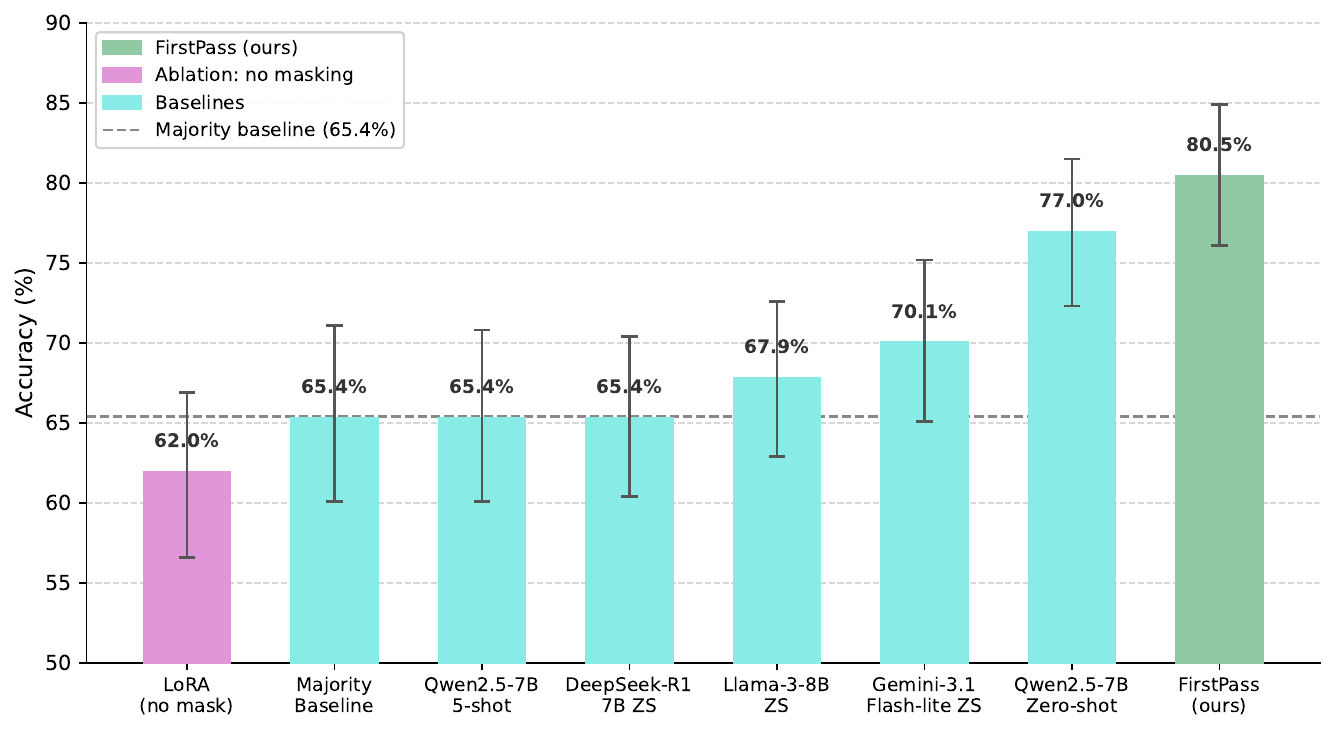}
\caption{Revision-cycle prediction accuracy with 95\% bootstrap confidence intervals. \textsc{FirstPass} achieves 80.5\%, outperforming all baselines. The no-masking ablation (62.0\%) falls below the majority baseline, demonstrating that response-only loss masking is an architectural prerequisite.}
\label{fig:accuracy}
\end{figure}

\textsc{FirstPass} achieves 80.5\% accuracy (+15.1 pp over majority, +10.4 pp over Gemini), as visualised with 95\% bootstrap confidence intervals in Figure~\ref{fig:accuracy}, with McNemar $p < 0.001$ against all baselines except Qwen zero-shot ($p = 0.185$). This non-significance indicates fine-tuning refines an already-capable foundation rather than correcting fundamental failures. Three baselines (Qwen 5-shot, DeepSeek, majority) collapse to 65.4\%, predicting \textsc{Standard} universally, confirming that few-shot prompting and zero-shot reasoning models fail to engage reliably with long scientific contexts. The no-masking ablation (62.0\%) is the starkest result: fine-tuning without response masking actively destroys performance, yielding a model worse than majority-class prediction.

\textbf{Per-domain results.} Biology 83.8\% ($n=68$), physics 81.8\% ($n=66$), neuroscience 81.5\% ($n=65$), chemistry 77.6\% ($n=67$), earth science 76.9\% ($n=52$). The narrow 6.9 pp spread confirms generalization across disciplines: revision-cycle prediction captures structural signals of unresolved concerns rather than domain-specific vocabulary. Full per-domain breakdown with confidence intervals is reported in Appendix~\ref{app:perdomain} (Table~\ref{tab:perdomain}).

\subsection{Review Generation}

We evaluate five systems on 372 test examples (Table~\ref{tab:gen}). Human references average 2,155 words. ROUGE-1/2/L with 95\% bootstrap CIs; paired bootstrap significance on ROUGE-L.

\definecolor{firstpassgray}{RGB}{20, 100, 145}
\definecolor{headerblue}{RGB}{219, 234, 254}

\begin{table}[t]
\centering
\resizebox{\columnwidth}{!}{%
\begin{tabular}{>{\raggedright\arraybackslash}p{3.8cm} ccccc}
\toprule
\rowcolor{headerblue}
\textbf{Model} & \textbf{R-1} & \textbf{R-2} & \textbf{R-L} & \textbf{Len} & \textbf{$p$} \\
\midrule
Qwen2.5-7B zero-shot     & 0.373 & 0.090 & 0.136 & 1,019 & $<$0.001 \\[2pt]
Llama-3-8B zero-shot     & 0.391 & 0.102 & $0.164^{\ddagger}$ & 1,006 & $<$0.001 \\[2pt]
DeepSeek-R1-7B zero-shot & 0.277 & 0.068 & 0.116 & 538   & $<$0.001 \\[2pt]
Gemini-3.1-flash-lite-preview ZS & 0.399 & 0.109 & 0.146 & 934   & $<$0.001 \\[2pt]
\textcolor{firstpassgray}{\textbf{\textsc{FirstPass} SFT (ours)}}
  & \textcolor{firstpassgray}{\textbf{0.354}}
  & \textcolor{firstpassgray}{\textbf{0.088}}
  & \textcolor{firstpassgray}{\textbf{0.154}}
  & \textcolor{firstpassgray}{\textbf{1,187}}
  & \textcolor{firstpassgray}{\textbf{---}} \\[2pt]
\midrule
Human reference          & ---   & ---   & ---   & 2,155 & --- \\
\bottomrule
\end{tabular}}
\vspace{10pt}
\caption{Review generation results ($n=372$). Human reference avg: 2,155 words. Paired bootstrap $p$ vs.\ \textsc{FirstPass} on ROUGE-L. $^\ddagger$Higher ROUGE-L than \textsc{FirstPass} ($\Delta = -0.009$, $p < 0.001$); reflects length artifact, not quality (see \S5.2).}
\label{tab:gen}
\end{table}

\textsc{FirstPass} achieves ROUGE-L 0.154, significantly outperforming Qwen ($\Delta = +0.018$), DeepSeek ($\Delta = +0.039$), and Gemini ($\Delta = +0.008$), all $p < 0.001$. Llama achieves the highest absolute ROUGE-L (0.164), but this is artifactual: Llama generates tightly constrained outputs (${\sim}$1,006 words) converging on common review phrases, optimizing n-gram overlap at the cost of depth. \textsc{FirstPass} produces longer reviews (1,187 words, closest to the human reference of 2,155) with lower TTR (0.212 vs.\ Llama's 0.229), consistent with expert \textit{Nature Communications} reviews that are structurally repetitive but content-rich.

\textbf{Per-domain ROUGE-L.} Chemistry 0.161, physics 0.159, neuroscience 0.158, biology 0.149, earth science 0.146. Consistent pattern confirms cross-domain capability rather than single-discipline overfitting. Complete generation metrics with bootstrap CIs for all models are provided in Appendix~\ref{app:genmetrics} (Table~\ref{tab:fullgen}).

\section{Discussion}

\textbf{\textsc{FirstPass} as scientific co-author.}
The workshop asks whether AI systems are tools, co-authors, or founders. \textsc{FirstPass} provides empirical evidence grounded in real editorial outcomes across five scientific disciplines. A tool assists without exercising judgment. A co-author tells you, before you submit, which concerns will survive peer review and whether your manuscript will require a second revision cycle. \textsc{FirstPass} does exactly this: it predicts extended revision cycles with 80.5\% accuracy and generates expert-length reviews identifying specific methodological weaknesses by section. Authors who use it in the pre-submission loop arrive at submission with stronger manuscripts, shorter rebuttals, and fewer unresolved Round~2 concerns. Consistent performance across biology, chemistry, neuroscience, physics, and earth science (76.9\% to 83.8\%) confirms that this judgment captures domain-general signals of unresolved reviewer concern, which is precisely the property a scientific co-author must have to be trusted across fields. 

\textbf{On the Qwen zero-shot result.} Qwen2.5-7B zero-shot achieves F1-macro 73.3\%, not significantly different from \textsc{FirstPass} ($p = 0.185$). This does not undermine fine-tuning: the base model already encodes strong scientific priors, and \textsc{FirstPass} refines rather than corrects it. The critical distinction is deployability: zero-shot Qwen requires the full 10,000-token peer review dialogue at inference time and produces no outcome prediction. \textsc{FirstPass}, by contrast, is a self-contained fine-tuned system that predicts revision cycles and generates domain-appropriate reviews from a single locally-deployable 7B model.

\textbf{The masking finding generalizes.} The 18.5 pp swing between masked and unmasked LoRA is not a peer-review curiosity: it is a general warning for any long-input/short-output classification task. When inputs dwarf targets, standard full-sequence loss training is actively harmful: gradient signal from thousands of input tokens drowns the classification objective entirely. Response-only masking is the correct default for this regime; \textsc{FirstPass} provides the first controlled empirical demonstration at scale.

\textbf{What 80.5\% accuracy means.} Four in five manuscripts correctly classified. For a journal receiving thousands of submissions annually, this is a meaningful productivity signal: \textsc{Extended} predictions flag papers needing senior reviewer assignment or closer editorial monitoring from Round 1. Consistent per-domain performance (76.9\% to 83.8\%) confirms deployability across multidisciplinary journals without per-domain recalibration.

\textbf{ROUGE is a floor, not a ceiling.} Llama's higher ROUGE-L reflects short, formulaic outputs optimizing n-gram overlap, not superior quality. \textsc{FirstPass} generates reviews closer to human length (1,187 vs.\ 2,155 words, vs.\ Llama's 1,006) with section-specific critique and technical depth. Human evaluation by domain scientists remains the gold standard.

\textbf{Limitations.} Five bounds: (1) only published papers available: rejected manuscripts, the strongest signal about publishability boundaries, are inaccessible. Extension to acceptance prediction, if rejected manuscripts become accessible through author-consent pipelines, is a direct next step that would sharpen the co-authorship claim by moving the evaluation from revision-cycle severity to publishability itself. (2) figures and supplementary data drive substantial reviewer concerns but are absent from our text pipeline, and multimodal extension via vision-language models is a natural next step; (3) \textsc{Standard}/\textsc{Extended} is a round-count proxy, not a direct measurement of scientific quality or concern severity; extra rounds can arise from reviewer communication style, editorial logistics, or field-specific norms rather than unresolved methodological debt. A targeted human annotation study confirming that \textsc{Extended} examples contain systematically more serious unresolved concerns would sharpen this label's validity. (3b) The model may exploit shortcut phrases in reviewer dialogue, such as ``major revision required'' or repeated concern restatements, that surface-signal the outcome without capturing deep scientific judgment; input-stage ablations isolating manuscript-only versus full-dialogue performance are a direct diagnostic. (4) the co-authorship use case is presented as a deployment framing rather than a validated workflow: our evaluation measures prediction accuracy on completed dialogues, and a prospective validation study, in which authors use FIRSTPASS before submission and subsequent revision outcomes are tracked against predictions, is the most direct path to validating the co-authorship claim and remains the immediate next step. (5) generalizability beyond \textit{Nature Communications}  remains untested: NC's mandatory transparent review, dense review format (avg 2,155 words), and multidisciplinary editorial norms may not transfer directly to outlets such as \textit{eLife} or \textit{PLOS ONE}, and cross-journal validation is a natural extension that would establish whether FIRSTPASS captures universal signals of scientific judgment or journal-specific editorial culture.

\section{Conclusion}

We presented \textsc{FirstPass}, the first multi-domain, multi-round peer review dataset and model grounded in real editorial outcomes. Three findings stand out: (1) response-only loss masking is a prerequisite for long-input scientific classification: omitting it collapses performance below the majority baseline, applying it yields 80.5\% accuracy and F1-macro 78.2\%; (2) outcome-grounded evaluation reveals what stylistic metrics cannot: models that look like good reviewers are not necessarily producing better scientific critique; (3) a 7B open-weight model achieves consistent cross-domain performance across five scientific disciplines, confirming that revision-cycle prediction captures domain-general signals of unresolved reviewer concern. Positioned at the boundary between tool and co-author on the spectrum this workshop examines, \textsc{FirstPass} demonstrates that a 7B open-weight model, trained on real multi-round editorial dialogues and evaluated against real outcomes, can exercise the anticipatory scientific judgment that authors need before submission and that the field needs to measure before deploying AI in scientific governance.

\clearpage


\bibliography{example_paper}
\bibliographystyle{icml2026}

\clearpage
\appendix

\section{System Prompts}
\label{app:prompts}

\textbf{Classification (all inference models).}
\begin{quote}
\textit{You are a senior editor at Nature Communications with deep expertise across biology, chemistry, earth science, neuroscience, and physics. Based on the peer review dialogue provided, predict the editorial outcome. Consider: severity of unresolved methodological concerns, number of outstanding reviewer requests, whether authors adequately addressed core issues, and the overall trajectory of the review dialogue. Answer with exactly one word on the last line: STANDARD or EXTENDED.}
\end{quote}

\textbf{Review generation (all inference models).}
\begin{quote}
\textit{You are assisting with a research study on automated scientific peer review. The following is an excerpt from a manuscript submitted to Nature Communications. As part of this NLP evaluation study, write the kind of detailed expert peer review that would appear in Nature Communications transparent peer review files. Your review must cover: 
(1) significance and novelty of the scientific contribution, 
(2) soundness of the methodology and experimental design, 
(3) quality and reproducibility of the results, 
(4) clarity and completeness of the reporting,
(5) statistical rigor where applicable,
(6) specific weaknesses that must be addressed before publication. 
Be specific - cite section names and claims where relevant. Write the full review text directly, without preamble.}
\end{quote}

\textbf{CLS training.}
\begin{quote}
\textit{You are a senior editor at Nature Communications with deep expertise across biology, chemistry, earth science, neuroscience, and physics. Based on the peer review dialogue provided, predict the editorial outcome. Consider: severity of unresolved methodological concerns, number of outstanding reviewer requests, whether authors adequately addressed core issues, and the overall trajectory of the review dialogue. Answer with exactly one word on the last line: STANDARD or EXTENDED.}
\end{quote}

\textbf{SFT training.} Review generation: \textit{You are a rigorous, constructive expert reviewer for Nature Communications. Write a detailed peer review covering significance, methodology, statistical rigor, and specific weaknesses.} Reviewer updating: \textit{You are a peer reviewer for Nature Communications writing a Round 2 review. Evaluate whether the authors satisfactorily addressed your Round 1 concerns.}

\section{Dataset Statistics}
\label{app:stats}

\definecolor{firstpassgray}{RGB}{20, 100, 145}
\definecolor{headerblue}{RGB}{219, 234, 254}
\definecolor{totalrow}{RGB}{240, 245, 252}

\begin{table}[h]
\centering
\resizebox{\columnwidth}{!}{%
\begin{tabular}{>{\raggedright\arraybackslash}p{2.2cm} ccccccc}
\toprule
\rowcolor{headerblue}
\textbf{Domain} & \textbf{Total} & \textbf{Train} & \textbf{Val} & \textbf{Test} & \textbf{STD (\%)} & \textbf{EXT (\%)} & \textbf{Avg Review (w)} \\
\midrule
Biology      & 741 & 593 & 74 & 74 & 63.2 & 36.8 & 2,201 \\[2pt]
Chemistry    & 744 & 595 & 75 & 74 & 62.8 & 37.2 & 2,183 \\[2pt]
Neuroscience & 739 & 591 & 74 & 74 & 64.6 & 35.4 & 2,144 \\[2pt]
Physics      & 727 & 582 & 73 & 72 & 66.7 & 33.3 & 2,098 \\[2pt]
Earth Sci.   & 717 & 573 & 72 & 72 & 71.2 & 28.8 & 2,149 \\[2pt]
\midrule
\rowcolor{totalrow}
\textcolor{firstpassgray}{\textbf{Total}}
  & \textcolor{firstpassgray}{\textbf{3,668}}
  & \textcolor{firstpassgray}{\textbf{2,934}}
  & \textcolor{firstpassgray}{\textbf{368}}
  & \textcolor{firstpassgray}{\textbf{366}}
  & \textcolor{firstpassgray}{\textbf{65.4}}
  & \textcolor{firstpassgray}{\textbf{34.6}}
  & \textcolor{firstpassgray}{\textbf{2,155}} \\
\bottomrule
\end{tabular}}
\vspace{10pt}
\caption{Dataset statistics by domain. STD = \textsc{Standard} (2-rounds), EXT = \textsc{Extended} (3+ rounds). Split: 80/10/10 at paper level, stratified by domain and label.}
\label{tab:datastats}
\end{table}

Table~\ref{tab:datastats} reports the full domain-level breakdown of the \textsc{FirstPass} dataset. The class balance is consistent across domains (62.8\% to 71.2\% \textsc{Standard}), confirming that the stratified split maintains representative label distributions in each domain. Earth Science has the highest \textsc{Standard} rate (71.2\%), suggesting methodological concerns in that domain are more frequently resolved within two rounds. 

\section{Truncation and Input Length Analysis}
\label{app:truncation}

Input length distributions vary substantially across tasks. Outcome prediction: median 9,847 tokens (mean 11,203, max 31,441), 34.2\% truncated at the 12,288-token limit. Review generation: median 4,312 tokens (mean 5,108, max 18,934), 8.7\% truncated at the 16,384-token limit. Symmetric 55/45 truncation retains the paper abstract, introduction, methods, and early results in the head segment, and the most recent reviewer exchange in the tail segment, ensuring both scientific content and the latest dialogue state are always visible to the model regardless of truncation.

\section{Per-Domain Classification Results}
\label{app:perdomain}

\definecolor{firstpassgray}{RGB}{20, 100, 145}
\definecolor{headerblue}{RGB}{219, 234, 254}

\begin{table}[h]
\centering
\resizebox{\columnwidth}{!}{%
\begin{tabular}{>{\raggedright\arraybackslash}p{2.2cm} cccc}
\toprule
\rowcolor{headerblue}
\textbf{Domain} & \textbf{n} & \textbf{Acc (\%)} & \textbf{F1-mac (\%)} & \textbf{F1-EXT (\%)} \\
\midrule
Biology       & 68 & 83.8 \scriptsize{[75.0, 91.2]} & 82.5 & 77.6 \\[2pt]
Chemistry     & 67 & 77.6 \scriptsize{[68.7, 88.1]} & 75.9 & 69.4 \\[2pt]
Neuroscience  & 65 & 81.5 \scriptsize{[72.3, 90.8]} & 78.9 & 71.4 \\[2pt]
Physics       & 66 & 81.8 \scriptsize{[72.7, 90.9]} & 80.0 & 73.9 \\[2pt]
Earth Science & 52 & 76.9 \scriptsize{[65.4, 86.5]} & 70.7 & 57.1 \\
\bottomrule
\end{tabular}}
\vspace{10pt}
\caption{\textsc{FirstPass} per-domain classification performance. Bootstrap 95\% CIs in brackets.}
\label{tab:perdomain}
\end{table}

Table~\ref{tab:perdomain} reports \textsc{FirstPass} classification performance broken down by scientific domain. The 6.9 pp spread between the best (Biology, 83.8\%) and worst (Earth Science, 76.9\%) domain confirms consistent generalization. F1-EXT drops most sharply in Earth Science (57.1\%), reflecting two compounding factors: the smaller number of EXTENDED examples ($n=15$) in that domain's test split inflates minority-class variance, and earth science spans internally heterogeneous subfields (geology, climatology, atmospheric science, oceanography) that a single domain label cannot fully condition on. Fine-grained subdomain conditioning is a direct improvement path.

\section{Full Generation Metrics}
\label{app:genmetrics}

\definecolor{firstpassgray}{RGB}{20, 100, 145}
\definecolor{headerblue}{RGB}{219, 234, 254}

\begin{table}[h]
\centering
\resizebox{\columnwidth}{!}{%
\begin{tabular}{>{\raggedright\arraybackslash}p{2.2cm} cccc}
\toprule
\rowcolor{headerblue}
\textbf{Model} & \textbf{R-1} & \textbf{R-2} & \textbf{R-L} & \textbf{TTR} \\
\midrule
Qwen ZS     & 0.373 \scriptsize{[0.362, 0.384]} & 0.090 \scriptsize{[0.086, 0.094]} & 0.136 \scriptsize{[0.132, 0.140]} & 0.367 \\[2pt]
Llama ZS    & 0.391 \scriptsize{[0.380, 0.401]} & 0.102 \scriptsize{[0.098, 0.106]} & 0.164 \scriptsize{[0.159, 0.168]} & 0.229 \\[2pt]
DeepSeek ZS & 0.277 \scriptsize{[0.267, 0.287]} & 0.068 \scriptsize{[0.064, 0.072]} & 0.116 \scriptsize{[0.112, 0.120]} & 0.417 \\[2pt]
Gemini ZS   & 0.399 \scriptsize{[0.389, 0.410]} & 0.109 \scriptsize{[0.104, 0.113]} & 0.146 \scriptsize{[0.142, 0.151]} & 0.397 \\[2pt]
\textcolor{firstpassgray}{\textbf{FirstPass}}
  & \textcolor{firstpassgray}{\textbf{0.354}} \scriptsize{[0.343, 0.364]}
  & \textcolor{firstpassgray}{\textbf{0.088}} \scriptsize{[0.084, 0.091]}
  & \textcolor{firstpassgray}{\textbf{0.154}} \scriptsize{[0.151, 0.158]}
  & \textcolor{firstpassgray}{\textbf{0.212}} \\
\bottomrule
\end{tabular}}
\vspace{10pt}
\caption{Per-model generation metrics with 95\% bootstrap CIs.}
\label{tab:fullgen}
\end{table}

Table~\ref{tab:fullgen} provides complete ROUGE scores with 95\% bootstrap confidence intervals for all five generation models. The non-overlapping CIs between \textsc{FirstPass} and DeepSeek on ROUGE-L confirm the significance of that comparison. Gemini's low TTR variance (0.397 vs.\ \textsc{FirstPass}'s 0.212) reflects its internal length regularization producing highly consistent but shorter outputs.

\section{McNemar Test Contingency Tables}
\label{app:mcnemar}

\definecolor{firstpassgray}{RGB}{20, 100, 145}
\definecolor{headerblue}{RGB}{219, 234, 254}
\definecolor{nsrow}{RGB}{254, 243, 199}

\begin{table}[h]
\centering
\footnotesize 
\begin{tabular}{>{\raggedright\arraybackslash}p{2.5cm} ccc}
\toprule
\rowcolor{headerblue}
\textbf{Baseline} & \textbf{$b$} & \textbf{$c$} & \textbf{$p$-value} \\
\midrule
Majority     & 28 & 76 & $4.0 \times 10^{-6}$ \\[2pt]
Qwen ZS      & 23 & 34 & $0.185^{\dagger}$ \\[2pt]
Qwen 5-shot  & 28 & 76 & $4.0 \times 10^{-6}$ \\[2pt]
Llama ZS     & 26 & 66 & $4.8 \times 10^{-5}$ \\[2pt]
DeepSeek ZS  & 28 & 76 & $4.0 \times 10^{-6}$ \\[2pt]
Gemini ZS    & 26 & 59 & $5.2 \times 10^{-4}$ \\[2pt]
LoRA no-mask & 18 & 77 & $\approx 0$ \\
\bottomrule
\end{tabular}
\vspace{10pt} 
\caption{McNemar's exact test vs.\ \textsc{FirstPass}. $b$ = baseline correct and \textsc{FirstPass} wrong; $c$ = \textsc{FirstPass} correct and baseline wrong. $^\dagger$Not significant: Qwen zero-shot and \textsc{FirstPass} make similar errors.}
\label{tab:mcnemar}
\end{table}

Table~\ref{tab:mcnemar} reports the full McNemar contingency values underlying the significance tests in Section~5.1. The $c$ column, cases where \textsc{FirstPass} is correct and the baseline is wrong, consistently exceeds $b$ for all baselines except Qwen zero-shot, confirming that \textsc{FirstPass}'s improvements are not due to trading one error type for another but represent genuine gains across both label classes.

\section{Qualitative Generation Examples}
\label{app:qualitative}

\textbf{Zero-shot Qwen ($\sim$1,019 words):} Generic structure, repetitive phrasing (``the paper is well written''), vague concerns (``methodology could be improved''), no specific section citations.

\textbf{\textsc{FirstPass} SFT ($\sim$1,187 words):} Specific section references (``the Gaussian approximation in Section 3.2''), reviewer dialogue awareness (``the rebuttal fails to address Reviewer 2's concern regarding contamination of the control group''), appropriate technical depth for domain.

\end{document}